\documentclass[10pt,twocolumn,letterpaper]{article}
\def\confName{WACV}
\def\confYear{2026}
\usepackage[applications]{wacv} 

\usepackage{algorithm}
\usepackage{algpseudocode}
\usepackage{amsmath}
\usepackage{multirow}
\usepackage{xcolor}
\usepackage{dsfont}
\usepackage{float} 
\usepackage{booktabs}
\usepackage{amssymb}
\usepackage{amsmath}
\usepackage{amsthm}
\usepackage{lipsum}
\usepackage{wrapfig} 
\usepackage{graphicx} 

\usepackage{pifont}

\usepackage[normalem]{ulem} 
\usepackage{colortbl} 
\usepackage{soul}

\definecolor{lightgray}{HTML}{FDEAE7}
\definecolor{wacvblue}{rgb}{0.21,0.49,0.74}
\usepackage[pagebackref,breaklinks,colorlinks,allcolors=wacvblue]{hyperref}

\author{
Abhinav Attri\thanks{Equal contribution} ,
Rajeev Ranjan Dwivedi\footnotemark[1],
Samiran Das,
and Vinod Kumar Kurmi\\
Indian Institute of Science Education and Research Bhopal, India\\
{\tt\small \{abhinav21, rajeev22, samiran, vinodkk\}@iiserb.ac.in}
}

\begin{document}

\title{Histogram Assisted Quality Aware Generative Model for Resolution Invariant NIR Image Colorization}

\maketitle

\begin{abstract}
We present HAQAGen, a unified generative model for resolution-invariant NIR-to-RGB colorization that balances chromatic realism with structural fidelity. The proposed model introduces (i) a combined loss term aligning the global color statistics through differentiable histogram matching, perceptual image quality measure, and feature-based similarity to preserve texture information, (ii) local hue–saturation priors injected via Spatially Adaptive Denormalization (SPADE) to stabilize chromatic reconstruction, and (iii) texture-aware supervision within a Mamba backbone to preserve fine details. We introduce an adaptive-resolution inference engine that further enables high-resolution translation without sacrificing quality. Our proposed NIR-to-RGB translation model simultaneously enforces global color statistics and local chromatic consistency, while scaling to native resolutions without compromising texture fidelity or generalization. Extensive evaluations on FANVID, OMSIV, VCIP2020, and RGB2NIR using different evaluation metrics demonstrate consistent improvements over state-of-the-art baseline methods. HAQAGen produces images with sharper textures, natural colors, attaining significant gains as per perceptual metrics. These results position HAQAGen as a scalable and effective solution for NIR-to-RGB translation across diverse imaging scenarios.
\end{abstract}

\section{Introduction}
Near-Infrared (NIR) imaging unveils a hidden world beyond human perception, capturing important visual information beyond the visible region, recording image information from 780 nm to 1000 nm. This capability makes NIR imaging indispensable in domains such as surveillance, night vision applications \cite{niu2023nir}, where it pierces through darkness; autonomous driving \cite{li2021low, liu2021benchmarking}, where it enhances visibility in adverse conditions \cite{choi2015referenceless}. Compared to visual images, NIR images substantially reduce the scattering of small and micro-scale particles present in smoke and fog, increasing the visibility. Despite the enormous potential, applications of NIR imaging systems are somewhat limited because the human visual system, trained to analyze visual information, cannot comprehend raw infrared images. The NIR-to-RGB translation approach bridges this gap by generating vivid, colorized images aligned with human perception. However, the process is fraught with challenges stemming from the spectral and geometric disparity between the NIR and RGB images, and the perceptual difference between these two modalities \cite{liang2021improved}. 

Current NIR-to-RGB translation approaches display several predicaments restricting their practical utility. Existing methods often suffer from textural information loss, geometric aberrations, color distortions, oversmoothing or blurring, and poor generalization. One pervasive issue is texture loss: many techniques generate outputs that lack the fine details present in the NIR input, resulting in blurred or oversmoothed images \cite{9412411, Chen:25, jung2021multispectral,app121910087}. Additionally, color distortions further complicate colorization, as the generated RGB images frequently display unnatural hues or inconsistent mappings \cite{qu2025near, son2017near, liang2021improved}, reducing their overall reliability. Moreover, the majority of existing models are \textit{constrained by fixed input and output sizes}, rendering them inflexible for real-world applications where image dimensions vary widely. Besides, the absence of a sufficient number of diverse, paired NIR-RGB dataset prevent the study of the generalizability of the methods. Compounding these issues, evaluations are typically confined to a single dataset, casting doubt on the generalizability of these methods across diverse scenarios. Finally, computational inefficiency limits their deployment in time-sensitive contexts.

To address these challenges, we propose the \textit{Histogram-Assisted  Quality Aware Generative Model } \textbf{(HAQAGen)}, a unified translation framework that (i) recovers and preserves fine-grained texture via a texture-aware generation module, (ii) enables the model to produce vivid, natural-coloured images through histogram-based priors, (iii) generalises reliably across multiple datasets, and (iv) supports adaptive-resolution inference for variable input sizes. Our framework unifies these elements into a single pipeline that achieves competitive perceptual quality and texture fidelity.

The remainder of this paper is organized as follows: Section~\ref{sec:related_work} reviews prior work in NIR-to-RGB translation, Section~\ref{sec:methodology} details proposed architecture and objectives, Sections~\ref{sec:experiments} and~\ref{sec:results} describe the experimental setup and results, and Section~\ref{sec:conclusion} concludes with limitations and future directions.


\section{Related Work}
\label{sec:related_work}

NIR to RGB translation has received increasing attention in recent years, driven by its importance in various applications \cite{hickman2020colour}, low-light enhancement, remote sensing, and surveillance applications \cite{hickman2020colour}. Early approaches relied on handcrafted features and classical regression models that attempted to directly map NIR pixel intensities to RGB values \cite{zhang2022review}. These conceptually simple methods lacked robustness to complex scene variations and failed to scale beyond narrow domains, highlighting the need for deep learning solutions. Recent deep learning driven approaches, such as generative and transformer-based architectures emerged as dominant paradigms. We summarize prior works considering three key aspects: spectral translation frameworks, texture preservation mechanisms, and evaluation practices in the subsequent sections:

\smallskip
\textbf{GAN-based Models} The first wave of deep approaches applied adversarial learning to capture the complex mapping between NIR and RGB. Mehri \etal \cite{Mehri2019Colorizing} employed CycleGANs for unpaired spectral translation, enforcing cycle consistency and adversarial supervision to bridge the modality gap without paired data. \cite{babu2020pcsgan} utilized a combination of different loss functions in their GAN model. Yan et al. \cite{Yan2020MFF} considered multi-scale features in their GAN model, while the work \cite{suarez2017infrared} generated three noisy versions of the same NIR scene to allow the GAN model to robustly learn the features. Dou et al. \cite{dou2019asymmetric} introduced a cycle-GAN model that utilized distinct loss functions to improve robustness. While effective in principle, these models often suffer from unstable training and spectral ambiguity, leading to inconsistent colors.

\smallskip
\textbf{Transformer-based Models} Recently, researchers have turned to transformer-style models to exploit long-range dependencies to uncover the NIR-to-RGB mapping. The prominent \textit{ColorMamba}\cite{Zhai2024ColorMamb} \cite{Zhai2024ColorMamb} model augments a state-space transformer backbone with learnable padding tokens, local convolutional modules, and agent-based attention. The model produces sharp boundaries and improved spectral fidelity. In this work, we adopt ColorMamba modules within our backbone but extend them with dual-branch supervision and histogram-based priors. Yang \etal \cite{Yang2023bMPFNet} introduced a feature embedding strategy to better align statistical and semantic cues across modalities. The framework improves PSNR and structural similarity by embedding features at multiple resolutions. However, generalization across unseen domains remains a challenge for these models.

\smallskip
\textbf{Texture Preservation in Colorization}
The prevalent NIR image colorization methods are unable to retain fine-grained texture information since NIR images lead to geometric distortions. Besides, colorized images are easily degraded by oversmoothing. Among the works attempting to resolve this issue, Li \etal \cite{LIU202181} proposed a bi-stream texture-aware GAN that disentangles global structural cues from local details, fusing them to restore high-frequency components. Building on this, Yang \etal \cite{yang2021attention} introduced an attention-guided network with dedicated modules for semantic reasoning and texture transfer, combined through an adaptive fusion block. Although these advances highlight the necessity of texture retention, existing models are unable to preserve finer texture details and often lack scalability to diverse resolutions.

\smallskip
\textbf{Evaluation and Generalization}
Conventional metrics, such as PSNR and SSIM, for quantifying fidelity of the generated RGB images, generally measure the pixelwise similarity, rather than the do not perceptual fidelity.  To address this gap, Liu \etal \cite{rs15071829} developed a deep image quality assessment framework that jointly considers texture, contrast, and color realism. Besides, most models are benchmarked on a single dataset, raising concerns about domain overfitting. Yang \etal \cite{Yang2023MPFNet} reported how validating on multiple datasets leads to improved robustness, underscoring the importance of cross-domain generalization.

\smallskip
Although generative models, particularly GANs and transformer models utilizing domain-dependent loss functions, have advanced the state-of-the-art, three fundamental challenges remain. (i) retaining fine-grained texture fidelity on par with high-frequency fusion networks, (ii) generating realistically coloured images for both local and global regions, and (iii) scaling seamlessly to arbitrary resolutions while ensuring cross-dataset generalization. Our proposed framework resolves these predicaments by unifying diverse loss terms, enforcing retention of texture information, perceptual image quality, and histogram alignment, and introducing an efficient, adaptive-resolution engine capable of translating images/patches of varying shapes. To our knowledge, this is the first NIR-to-RGB system that simultaneously enforces global color statistics and local chromatic consistency, while scaling to native resolutions without compromising texture fidelity or generalization.

\section{Methodology}
\label{sec:methodology}

We envisage approaches for the translation of a single-channel NIR image
$\mathbf{x}_{\text{nir}}\in\mathbb{R}^{H\times W\times 1}$ to a three-channel RGB image
$\hat{\mathbf{y}}_{\text{rgb}}\in\mathbb{R}^{H\times W\times 3}$. Given paired supervision $(\mathbf{x}_{\text{nir}},\mathbf{y}_{\text{rgb}})$, we learn a mapping $\mathcal{F}_\Theta:\mathbf{x}_{\text{nir}}\mapsto\hat{\mathbf{y}}_{\text{rgb}}$ by minimizing a composite objective that balances (i) photometric and perceptual fidelity and (ii) chromatic realism under the inherent spectral ambiguity of NIR$\to$RGB. We denote the ground-truth HSV (hue/saturation/value) as $\mathbf{y}_{\text{hsv}}=\Psi(\mathbf{y}_{\text{rgb}})$ and the model’s auxiliary prediction as $\hat{\mathbf{y}}_{\text{hsv}}$. \textit{Colour-space conventions.} Unless stated otherwise, images are in sRGB and linearly scaled to $[0,1]$; HSV is computed from sRGB via $\Psi(\cdot)$ and used for auxiliary supervision and SPADE conditioning \cite{SPADE}. Losses that are defined “per channel” default to sRGB channels.
\begin{figure}[h]
    \centering
    \includegraphics[width=0.99\linewidth]{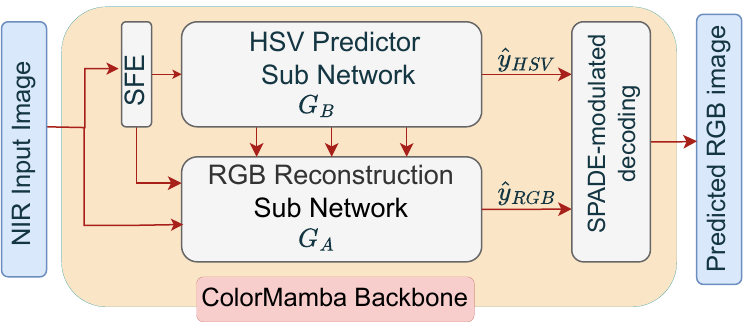} 
    \caption{Proposed framework. NIR features feed two branches: an HSV Predictor and an RGB Reconstruction network. HSV guides the RGB decoder via SPADE \cite{SPADE}, with dual discriminators and multi-term losses ensuring realism and consistency.}
    \label{fig:arch_overview}
\end{figure}

\subsection{Backbone and Overall Design}
\label{subsec:backbone}
\paragraph{Mamba-based encoder–decoder.}

We adopt \emph{ColorMamba}~\cite{Zhai2024ColorMamb} as the visual backbone for efficient representation learning while retaining long-range dependencies. Let $\mathcal{E}$ and $\mathcal{D}$ denote the shared encoder and decoder blocks, respectively.

\medskip
\noindent \textbf{Dual-branch generation with SPADE conditioning \cite{SPADE}.}
We introduce a dual generator $\mathcal{G}=\{G_A,G_B\}$:
$G_A$ is the \emph{RGB branch} that predicts $\hat{\mathbf{y}}_{\text{rgb}}$, and
$G_B$ is an \emph{HSV-prior branch} that regresses a dense hue–saturation–value field
$\hat{\mathbf{y}}_{\text{hsv}} = G_B(\mathbf{x}_{\text{nir}})$ from the same input.
To convey local chromatic priors into $G_A$, we inject $\hat{\mathbf{y}}_{\text{hsv}}$ into every decoder stage via SPADE-ResNet modulation~\cite{9301837}:
for a decoder feature map $\mathbf{F}\in\mathbb{R}^{h\times w\times c}$ we apply
\begin{equation}
\label{eq:spade}
\widehat{\mathbf{F}} \;=\; \gamma(\hat{\mathbf{y}}_{\text{hsv}})\odot \mathbf{F} \;+\; \beta_*(\hat{\mathbf{y}}_{\text{hsv}}),
\end{equation}
where $\gamma(\cdot),\beta_*(\cdot)$ are lightweight convolution blocks, and $\odot$ denotes Hadamard product.
Eq.~\eqref{eq:spade} equips $G_A$ with \emph{region-aware colour cues} while the backbone supplies geometry and texture.

\subsection{Learning Objective}
\label{subsec:loss}
Our objective comprises an adversarial tier that attempts to achieve naturalness in both colour spaces and a reconstruction tier that aligns texture, semantics, and global colour statistics.

\medskip
\noindent \textbf{Adversarial tier.}
Two PatchGAN discriminators $(D_{\text{RGB}},D_{\text{HSV}})$ operate on $\hat{\mathbf{y}}_{\text{rgb}}$ and $\hat{\mathbf{y}}_{\text{hsv}}$ respectively, enforcing complementary constraints on luminance/chrominance.
We use the \emph{hinge} adversarial loss with $70{\times}70$ receptive fields, spectral normalization on $D$, and a $1{:}1$ $G{:}D$ update ratio. For $c\!\in\!\{\text{RGB},\text{HSV}\}$,
{
\small
\begin{align}
\label{eq:adv_gen}
\mathcal{L}_{\text{GAN}}^{G,c} &= -\,\mathbb{E}\big[D_c(\hat{\mathbf{y}}_c)\big],\\
\label{eq:adv_disc}
\mathcal{L}_{\text{GAN}}^{D,c} &= \mathbb{E}\big[\max(0,1 - D_c(\mathbf{y}_c))\big] + \mathbb{E}\big[\max(0,1 + D_c(\hat{\mathbf{y}}_c))\big],\nonumber
\end{align}
}
and $\mathcal{L}_{\text{GAN}}^{G}=\sum_c \mathcal{L}_{\text{GAN}}^{G,c}$, $\mathcal{L}_{\text{GAN}}^{D}=\sum_c \mathcal{L}_{\text{GAN}}^{D,c}$.
Supervising RGB and HSV with distinct critics makes hue failures detectable even when luminance appears plausible.

\paragraph{HAQAGen reconstruction tier.}
We regularize with a multi-purpose feature- and statistics-aware loss

{
\small
\begin{equation}
\label{eq:lrec}
\begin{aligned}
\mathcal{L}_{\text{rec}}(\hat{\mathbf{y}},\mathbf{y})
&=
\underbrace{\alpha\;\|f(\hat{\mathbf{y}})-f(\mathbf{y})\|_2^2
+\gamma\bigl[1-\cos(f(\hat{\mathbf{y}}),f(\mathbf{y}))\bigr]}_{\text{task-specific texture basis (frozen autoencoder)}} \\
&\quad+ \underbrace{\beta\;\|\operatorname{CDF}(\hat{\mathbf{y}})-\operatorname{CDF}(\mathbf{y})\|_1}_{\text{global colour prior (differentiable CDF)}} \\
&\quad+ \underbrace{\delta\;\|g(\hat{\mathbf{y}})-g(\mathbf{y})\|_2^2}_{\text{perceptual mid-level semantics (VGG-19)}}
\end{aligned}
\end{equation}
}
with $(\alpha,\beta,\gamma,\delta)=(1.0,1.5,1.0,0.2)$.
Here $f(\cdot)$ is a frozen four-layer autoencoder capturing task-specific textural features, and $g(\cdot)$ extracts \texttt{relu4\_2} activations from VGG-19. The autoencoder terms stabilize high-frequency detail; the VGG term anchors semantic structure; the CDF term combats colour drift.

\medskip
\noindent \textbf{Differentiable histogram loss.}
Following~\cite{avi2023differentiable}, we compute a soft histogram $\mathbf{h}\in\mathbb{R}^{B}$ for each \emph{sRGB} channel with temperature $\tau$ and bin centers $\{c_b\}_{b=1}^B$:
$
h_b = \tfrac{1}{N}\sum_{i=1}^{N} k_\tau(\hat{y}_i-c_b),
$
where $k_\tau$ is a smooth kernel (e.g., triangular or logistic); the CDF is $\mathbf{H}$ with $H_b=\sum_{j\le b} h_j$.
We set $B{=}64$ and $\tau{=}0.02$ by default (see sensitivity in Sec.~\ref{sec:results}).
We attempt to penalize the mismatch between output and target CDFs channel-wise and average across channels using $\ell_1$ norm. The loss term yields stable, smooth gradients that align global chromatic statistics without distorting the local structure.

\medskip
\noindent \textbf{Full objective:}

{\small
\begin{equation}
\label{eq:gen_disc}
\begin{aligned}
\mathcal{L}_G
&=
\lambda_{\text{adv}}\,\mathcal{L}_{\text{GAN}}^{G}
+\lambda_{\text{mse}}\!\left[
\text{MSE}(\hat{\mathbf{y}}_{\text{rgb}},\mathbf{y}_{\text{rgb}})
+\text{MSE}(\hat{\mathbf{y}}_{\text{hsv}},\mathbf{y}_{\text{hsv}})
\right]
\\[-1pt]
&\qquad\qquad
+\lambda_{\text{feat}}\!\left[
\mathcal{L}_{\text{rec}}(\hat{\mathbf{y}}_{\text{rgb}},\mathbf{y}_{\text{rgb}})
+\mathcal{L}_{\text{rec}}(\hat{\mathbf{y}}_{\text{hsv}},\mathbf{y}_{\text{hsv}})
\right],
\\[4pt]
\mathcal{L}_D
&=
\mathcal{L}_{\text{GAN}}^{D}.
\end{aligned}
\end{equation}
}

\begin{algorithm}[]
\caption{Dynamic Patching – Sliding-Window Inference}
\label{alg:crisp_sliding_window}
\begin{algorithmic}[1]
\Require NIR image $\mathbf{I}$; model $M$; patch size $P$; overlap $O$
\Ensure RGB image $\mathbf{O}$
\State $S \gets P - O$ \Comment{stride}
\State Pad $\mathbf{I}$ to cover multiples of $S$
\State Compute grid $\{(y_i,x_i)\}_{i=1}^N$ of patch origins
\State Build feather mask $\mathbf{M} \in \mathbb{R}^{P\times P}$
\State Initialize accumulators $\mathbf{O}_{\mathrm{pad}}, \mathbf{W}_{\mathrm{pad}} \gets 0$
\For{$i=1\ldots N$}
  \State Extract patch $\mathbf{p}_i \gets \mathbf{I}[y_i:y_i+P,\;x_i:x_i+P]$
  \State Predict $\hat{\mathbf{y}}_i \gets M(\mathbf{p}_i)$
  \State Add $\hat{\mathbf{y}}_i \odot \mathbf{M}$ into $\mathbf{O}_{\mathrm{pad}}$
  \State Add $\mathbf{M}$ into $\mathbf{W}_{\mathrm{pad}}$
\EndFor
\State Normalize $\mathbf{O}_{\mathrm{pad}} \gets \mathbf{O}_{\mathrm{pad}} \oslash \mathbf{W}_{\mathrm{pad}}$
\State Crop to original size and \Return $\mathbf{O}$
\end{algorithmic}
\end{algorithm}

\medskip
\noindent \textbf{Texture-Aware Feature Enhancement}
\label{subsec:texture}
Rather than relying solely on pixel losses, we capture \emph{high-level, intermediate} representations that correlate with human sensitivity to edges and micro-texture.
The feature similarity loss $\mathcal{L}_{\text{rec}}$ divides $(\hat{\mathbf{y}},\mathbf{y})$ includes two complementary terms:
(i) a frozen autoencoder $f(\cdot)$ that captures task-specific fine structure via $\ell_2$ and cosine similarity; and
(ii) a VGG-19 encoder $g(\cdot)$ capturing mid-level semantics via $\ell_2$.
All feature losses operate on $256{\times}256$ patches during training for stability, and over full-resolution outputs at inference for fidelity.
\textit{Autoencoder pretraining.} The texture autoencoder is trained \emph{once} on the union of training splits (no test images) using an $\ell_2$ reconstruction objective on $256{\times}256$ patches; after convergence, it is frozen and reused across all experiments to avoid dataset-specific leakage. Inputs are scaled to $[0,1]$; Autoencoder feature vectors are instance-normalized before computing Eq.~\eqref{eq:lrec}.

\medskip
\noindent \textbf{Global - Local Colour Guidance}
\label{subsec:colour_guidance}
Since the NIR images generally lack chromatic cues, purely local criteria are insufficient.
To resolve this issue, our model melds:
(i) a \emph{global} differentiable CDF loss (Eq.~\ref{eq:lrec}) that \emph{aligns global colour statistics}, and
(ii) \emph{local} HSV priors injected through SPADE \cite{SPADE} (Eq.~\ref{eq:spade}) so that decoder features are modulated by spatially varying hue/saturation hints.
This combination ensures identical NIR intensities correspond to different materials (e.g., foliage vs.\ rock), yielding globally realistic and locally coherent colourization.
\begin{figure*}[h]
    \centering
    \includegraphics[width=0.95\linewidth]{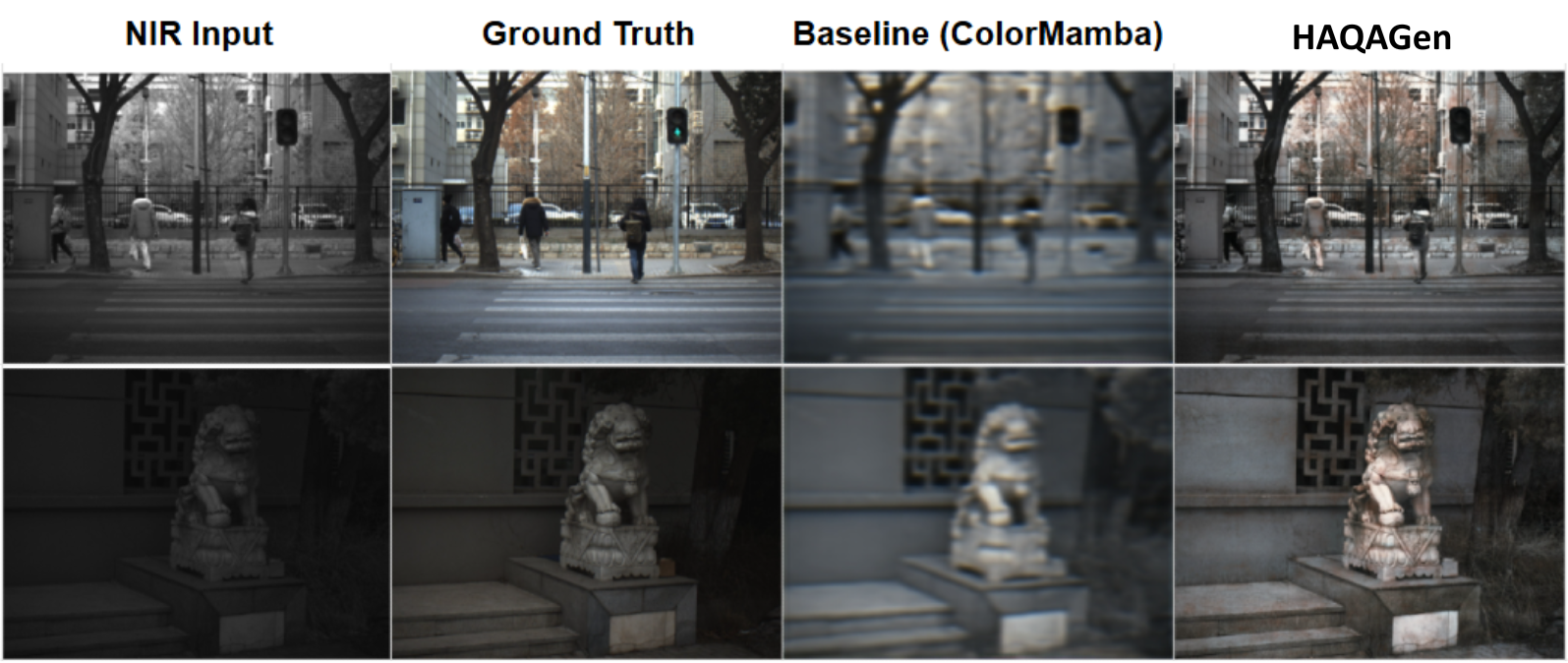 }
    \caption{Comparison of FANVID dataset: (1) NIR input, (2) ground-truth RGB, (3) prediction with resizing (blurred), (4) prediction with adaptive resolution (sharper texture, better color).}
    \label{fig:omsiv_patch_vs_resize}
\end{figure*}

\subsection{Adaptive-Resolution Inference}
\label{subsec:adaptive}
Since NIR images are generally high-resolution, na\"{i}ve resizing to $256\times 256$ introduces irreversible blur and texture loss. We therefore adopt a resolution-agnostic pipeline containing three components:

\medskip
\noindent \textbf{(i) Patch-based training.}
We optimize on overlapping $256{\times}256$ cropped patches to learn translation locally while regularizing with global terms (CDF/perceptual).

\medskip
\noindent \textbf{(ii) Sliding-window inference.}
At test time, we tile the image into overlapping patches of size $P=256$ with stride $S\in\{222,240\}$ (overlap $O=P{-}S$), process each patch independently with $(G_A, G_B)$, and recompose. We additionally report a sensitivity sweep over $S$ and a simple seam-energy diagnostic (gradient variance across patch borders) in Sec.~\ref{sec:results}.

\medskip
\noindent \textbf{(iii) Feather blending.}
Let $\mathbf{p}_i$ be the $i$-th output patch and $\mathbf{M}\in\mathbb{R}^{P\times P}$ a separable 2-D Hanning mask.
We accumulate
$\mathbf{O}_{\text{pad}} \mathrel{+}= \mathbf{p}_i\odot\mathbf{M}$ and the weights
$\mathbf{W}_{\text{pad}} \mathrel{+}= \mathbf{M}$, then normalize
$\mathbf{O} = \text{crop}\bigl(\mathbf{O}_{\text{pad}} \oslash \mathbf{W}_{\text{pad}}\bigr)$,
where $\oslash$ is element-wise division.
This eliminates seam artifacts and preserves edge continuity.

\begin{figure*}[h]
    \centering
    \includegraphics[width=.87\linewidth]{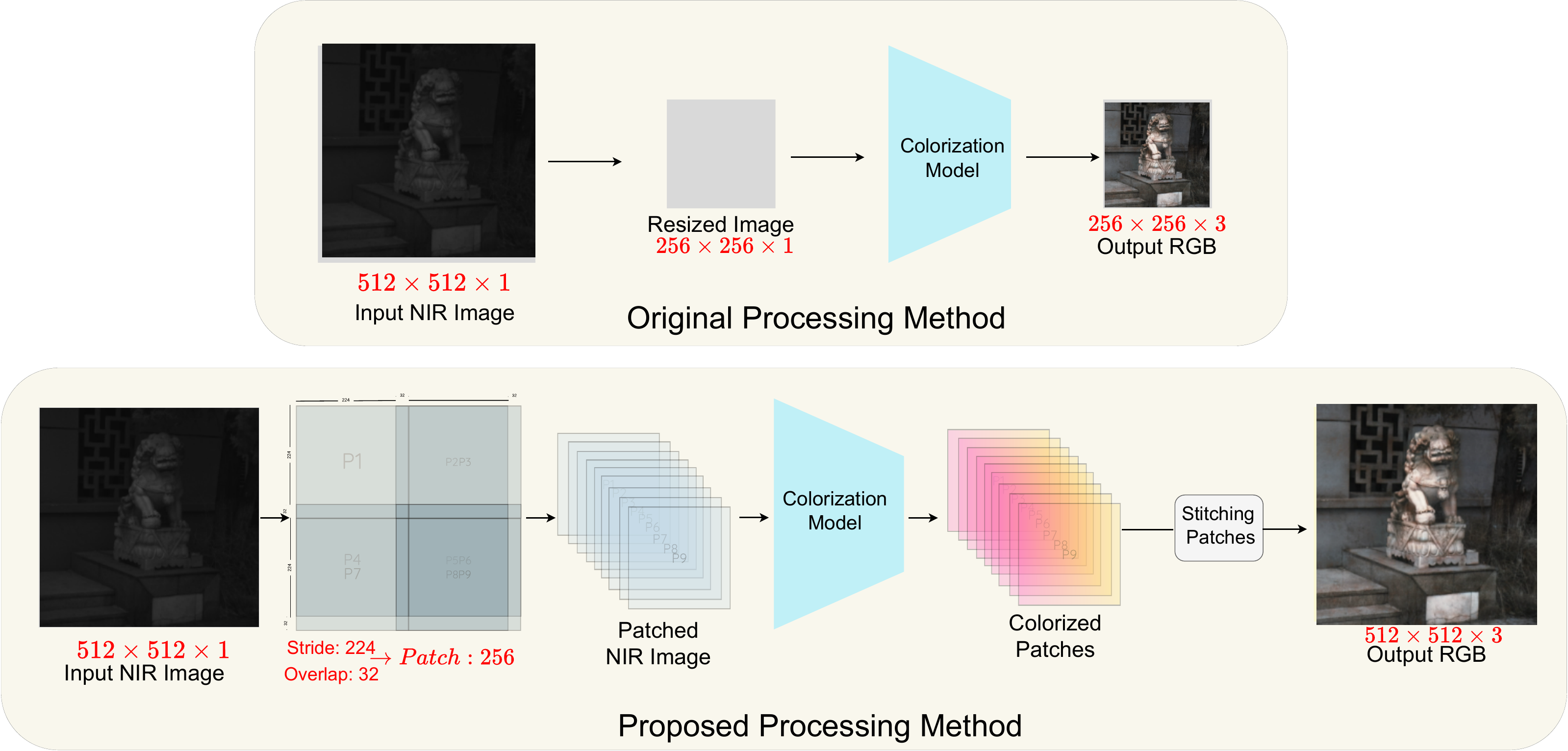} 
    \caption{Adaptive patching: stride-based tiling, patch-wise colorization, and feathered stitching for seamless RGB output.}
    \label{fig:patch_sweep}
\end{figure*}
\paragraph{Content-aware downscaling (mitigating uniform-patch bias).}
Large scenes often over-sample local regions that do not contain detailed texture information (e.g., sky), creating bias in training. We therefore isotropically clamp the long side to $\le 512$ during training-time cropping (preserving aspect ratio), improving semantic diversity of patches and stabilizing optimization.

\medskip
\noindent \textbf{Network Heads and Discriminators}
\label{subsec:heads}
\textbf{HSV prior branch ($G_B$).}
A compact depthwise-separable CNN (\,\,$\approx$4M params\,\,) regresses $\hat{\mathbf{y}}_{\text{hsv}}$.
\textbf{Texture encoder $f(\cdot)$.}
A frozen, lightweight 4-layer autoencoder (no skips) provides the texture basis in Eq.~\eqref{eq:lrec}. \textbf{Dual critics.} $D_{\text{RGB}}$ and $D_{\text{HSV}}$ are $70{\times}70$ PatchGANs sharing all weights except the first conv layer to respect channel semantics. All images are treated as sRGB in $[0,1]$; HSV targets are obtained via $\Psi(\cdot)$ from sRGB.

\section{Experimental Setup \& Implementation Details}
\label{sec:experiments}

\paragraph{Datasets}
\label{subsec:dataset}
We evaluate \textsc{HAQAGen} on four public benchmarks spanning faces, urban/outdoor scenes, and mixed environments:
FANVID~\cite{fanvid}, OMSIV~\cite{soria2017rgb-nirDataset}, VCIP2020~\cite{Yang2023aCoColor}, and RGB2NIR~\cite{BS11}.
FANVID contains 5{,}144 paired VIS--NIR images (700--800\,nm) at $2048{\times}1536$, emphasizing facial imagery and dynamic scenes.
OMSIV contains 532 NIR--RGB pairs at $580{\times}320$ covering varied outdoor settings.
VCIP2020 comprises 400 pairs at $256{\times}256$ across indoor/outdoor scenes.
RGB2NIR includes 477 TIFF image pairs with variable resolution (up to $1024{\times}768$) over nine categories (countryside, field, forest, indoor, mountain, old buildings, streets, urban, water).
Table~\ref{tab:dataset_stats} summarizes statistics and splits. Unless specified, we use official splits; otherwise, we adopt 80/10/10 train/val/test without scene overlap.

\paragraph{Training protocol.}
Unless stated otherwise, we train for $50$ epochs on random $256^2$ crops with AdamW ($\beta_1{=}0.5$, $\beta_2{=}0.999$, weight decay $10^{-4}$) and cosine learning rate decay ($1\times10^{-4}\!\to\!1\times10^{-6}$).
Training is performed in mixed precision (AMP, fp16) with a global batch size of $16$ across four RTX~4090 GPUs.
The composite objective weights are set to $\{\lambda_{\text{MSE}},\lambda_{\text{feat}},\lambda_{\text{adv}}\}{=}\,15{:}15{:}1$, balancing distortion, feature, and adversarial terms (see Sec.~\ref{sec:methodology}).
All hyperparameters match the implementation described in the manuscript, and normalizations and colour-space conversions follow Sec.~\ref{sec:experiments}.

\begin{table*}[t]
\centering
\small
\begin{tabular}{lcccccc}
\toprule
Dataset & Type & \#Pairs & Train / Val / Test & Modal Res. & Bit depth & Year \\
\midrule
VCIP2020 & indoor/outdoor & 400 & 320 / 40 / 40 & $256\times 256$ & 8 & 2020 \\
FANVID   & faces \& urban  & 5144 & 4100 / 514 / 530 & $2048\times1536$ & 8 & 2024 \\
OMSIV    & outdoor         & 532 & 426 / 53 / 53 & $580\times320$ & 8 & 2017 \\
RGB2NIR  & mixed scenes    & 477 & 382 / 48 / 47 & var.\ ($\le1024\times768$) & 16 & 2011 \\
\bottomrule
\end{tabular}
\caption{Dataset statistics and splits. Resolution reports the modal native size; “var.” indicates multiple aspect ratios.}
\label{tab:dataset_stats}
\end{table*}

\subsection{Preprocessing and Augmentation}
\label{subsec:prep}
We convert RGB images to \emph{linear} sRGB, then perform min-max normalization to convert both modalities to $[0,1]$. We also apply histogram equalization to the NIR channel to reduce illumination bias prior to training. Ground-truth RGB is converted to HSV (for colour-consistency checks), while predicted RGB is mapped to \mbox{sRGB$\!\to$LAB} for perceptual losses. Unless stated otherwise, quantitative metrics are computed on linear sRGB.

\noindent\textbf{Data augmentation.}
We employ random horizontal flips ($p{=}0.5$), $90^\circ$ rotations, and HSV-saturation jitter ($\pm$10\%) on the \emph{reference} RGB only.
For high-resolution datasets (FANVID, RGB2NIR), we additionally sample random $384{\times}384$ crops to encourage scale robustness prior to $256{\times}256$ patch formation.  

\noindent \textbf{Inference at Arbitrary Resolution}
\label{subsec:inference}
We adopt sliding-window inference with feather blending (Sec.~\ref{subsec:adaptive}) to avoid loss of detailed information due to na\"{i}ve resizing. We consider patch size $P{=}256$; stride $S\in\{222,240\}$ (overlap $16$--$34$\,px); Hanning feather masks for seamless stitching; reflective padding for small borders. The approach enables resolution-agnostic testing with preserved textures and clean seams.  
Representative qualitative examples displayed in Figs.~\ref{fig:qualitative} and \ref{fig:omsiv_resize} underlines that HAQAGen not only preserves the detailed information, but also matches the color, and geometric information.
 
\smallskip
\noindent \textbf{Evaluation Protocol and Metrics.} 
We evaluate our model using four complementary metrics. Peak Signal-to-Noise Ratio (PSNR)~\cite{hore2010image} measures pixelwise fidelity as $\mathrm{PSNR} = 10 \cdot \log_{10}\!\left(\tfrac{MAX_I^2}{\mathrm{MSE}}\right)$ with $\mathrm{MSE} = \tfrac{1}{N}\sum_i (x_i-y_i)^2$, where $MAX_I$ is the intensity range. Structural Similarity Index (SSIM)~\cite{hore2010image} captures luminance, contrast, and structural consistency via $\mathrm{SSIM}(x,y) = \tfrac{(2\mu_x\mu_y + C_1)(2\sigma_{xy}+C_2)}{(\mu_x^2+\mu_y^2+C_1)(\sigma_x^2+\sigma_y^2+C_2)}$, with $\mu$, $\sigma^2$ denoting local means and variances, and $\sigma_{xy}$ the covariance. Angular Error (AE)~\cite{androutsos1999novel} quantifies chromatic accuracy as $\mathrm{AE}(p,g) = \cos^{-1}\!\left(\tfrac{p \cdot g}{\|p\|\|g\|}\right)$, measuring hue differences while remaining invariant to intensity scaling. Finally, Learned Perceptual Image Patch Similarity (LPIPS)~\cite{lpips} estimates perceptual distance by comparing deep features, $\mathrm{LPIPS}(x,y) = \sum_l \tfrac{1}{H_lW_l}\sum_{h,w} \|\, w_l \odot (\phi_l(x)_{hw} - \phi_l(y)_{hw}) \,\|_2^2$, where $\phi_l$ are pretrained features and $w_l$ are learned channel weights. Together, PSNR and SSIM assess fidelity and structure, AE evaluates chromatic alignment, and LPIPS measures perceptual quality, holistically covering most perspectives holistically.

\medskip
\noindent \textbf{Baselines \& Reproduction}
\label{subsec:baselines}
We benchmark against three categories of baselines:  
\noindent (i) \emph{GAN-based} methods (e.g., CycleGAN~\cite{Mehri2019Colorizing}) trained with paired or unpaired NIR--RGB supervision;  
(ii) \emph{Transformer/ state-space models} such as ColorMamba~\cite{Zhai2024ColorMamb}, which we adopt as our backbone; and  
(iii) \emph{Texture-aware/attention models} (e.g., MPFNet~\cite{Yang2023bMPFNet}, AttentionGAN~\cite{yang2021attention}) that explicitly model texture or semantic priors.  

For fairness, all baselines are retrained (when open-source code is available) using our unified preprocessing pipeline and training schedule, with input resizing handled consistently: $256^2$ for VCIP2020/OMSIV, adaptive sliding-window for FANVID/RGB2NIR. When official pre-trained weights are used, we re-evaluate them under our metrics (PSNR, SSIM, AE, LPIPS) to ensure comparability.  
This harmonized protocol guarantees that reported gains stem from model design rather than differences in data processing or evaluation.  

\section{Results and Discussion}
\label{sec:results}

\begin{figure*}[htbp]
    \centering
    \includegraphics[width=0.88\linewidth]{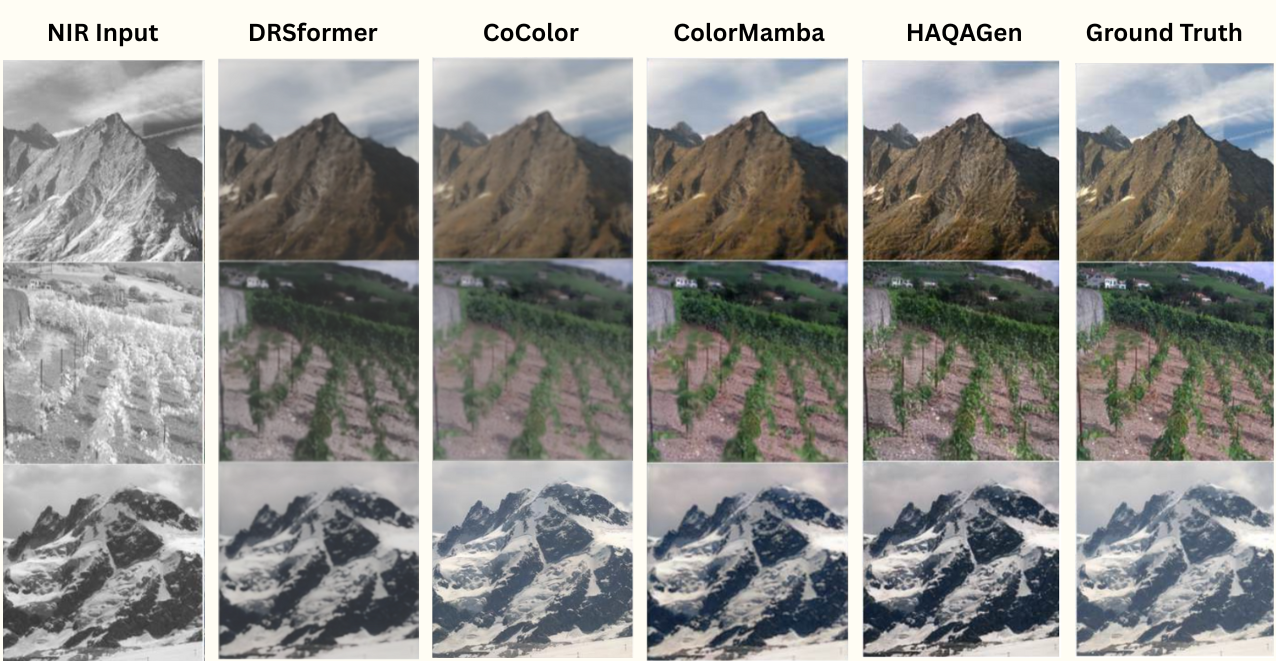}
    \caption{Qualitative comparison on the VCIP2020 dataset \cite{Yang2023aCoColor}: (1) NIR input, (2) DRSformer \cite{Chen2023DRSformer}, (3) CoColor \cite{Yang2023aCoColor}, (4) ColorMamba \cite{Zhai2024ColorMamb}, (5) our proposed HAQAGen, and (6) ground-truth RGB. HAQAGen achieves sharper textures, more natural chromatic distributions, and better structural fidelity compared to prior baselines.}
    \label{fig:qualitative}
\end{figure*}

\noindent
Table~\ref{tab:sota} benchmarks \textsc{HAQAGen} against twelve SOTA methods on VCIP2020. Our model achieves the best PSNR (24.96\,dB) and the lowest LPIPS (0.18), while matching the top SSIM (0.71). Although AE is marginally higher than ColorMamba (2.96 vs.\ 2.81), visual inspection in Fig.~\ref{fig:qualitative} indicates that this trade-off correlates with richer chroma and sharper textures. Across the broader set of baselines, \textsc{HAQAGen} reduces AE by at least 23.3\% (vs.\ SST) and LPIPS by 34.6\% (vs.\ NIR-GNN), indicating strong perceptual fidelity.

\begin{table}[h]
\centering 
\small
\caption{Quantitative results on VCIP2020. Best in \textbf{bold}.}
\scalebox{0.9}{
\begin{tabular}{lcccc}
\hline
\textbf{Methods} & \textbf{PSNR($\uparrow$)} & \textbf{SSIM($\uparrow$)} & \textbf{AE($\downarrow$)} & \textbf{LPIPS($\downarrow$)} \\
\hline
SST ~\cite{Yan2020MFF}           & 14.26          & 0.57          & 5.61          & 0.361 \\
NIR-GNN ~\cite{Valsesia2020NIRGNN}  & 17.50          & 0.60          & 5.22          & 0.384 \\
MFF ~\cite{Yan2020MFF}           & 17.39          & 0.61          & 4.69          & 0.318 \\
ATCGAN ~\cite{Yang2020CycleGAN}     & 19.59          & 0.59          & 4.33          & 0.295 \\
Restormer ~\cite{Zamir2022Restormer}   & 19.43          & 0.54          & 4.41          & 0.267 \\
DRSformer ~\cite{Chen2023DRSformer}   & 20.18          & 0.56          & 4.22          & 0.254 \\
MPFNet~\cite{Yang2023bMPFNet}     & 22.14          & 0.63          & 3.68          & 0.253 \\
CoColor ~\cite{Yang2023aCoColor}    & 23.54          & 0.69          & \textbf{2.68} & 0.233 \\
MCFNet ~\cite{Zhai2024MCFNet}       & 20.34          & 0.61          & 3.79          & 0.208 \\ 
ColorMamba~\cite{Zhai2024ColorMamb}    & 24.56 & \textbf{0.71} & 2.81    & 0.212 \\ 
\rowcolor{lightgray} \textsc{HAQAGen} & \textbf{24.96}  & \textbf{0.71}  & 2.96 & \textbf{0.18} \\
\hline
\end{tabular}}
\label{tab:sota}
\end{table}

\begin{figure*}
    \centering
    \includegraphics[width=0.9\linewidth]{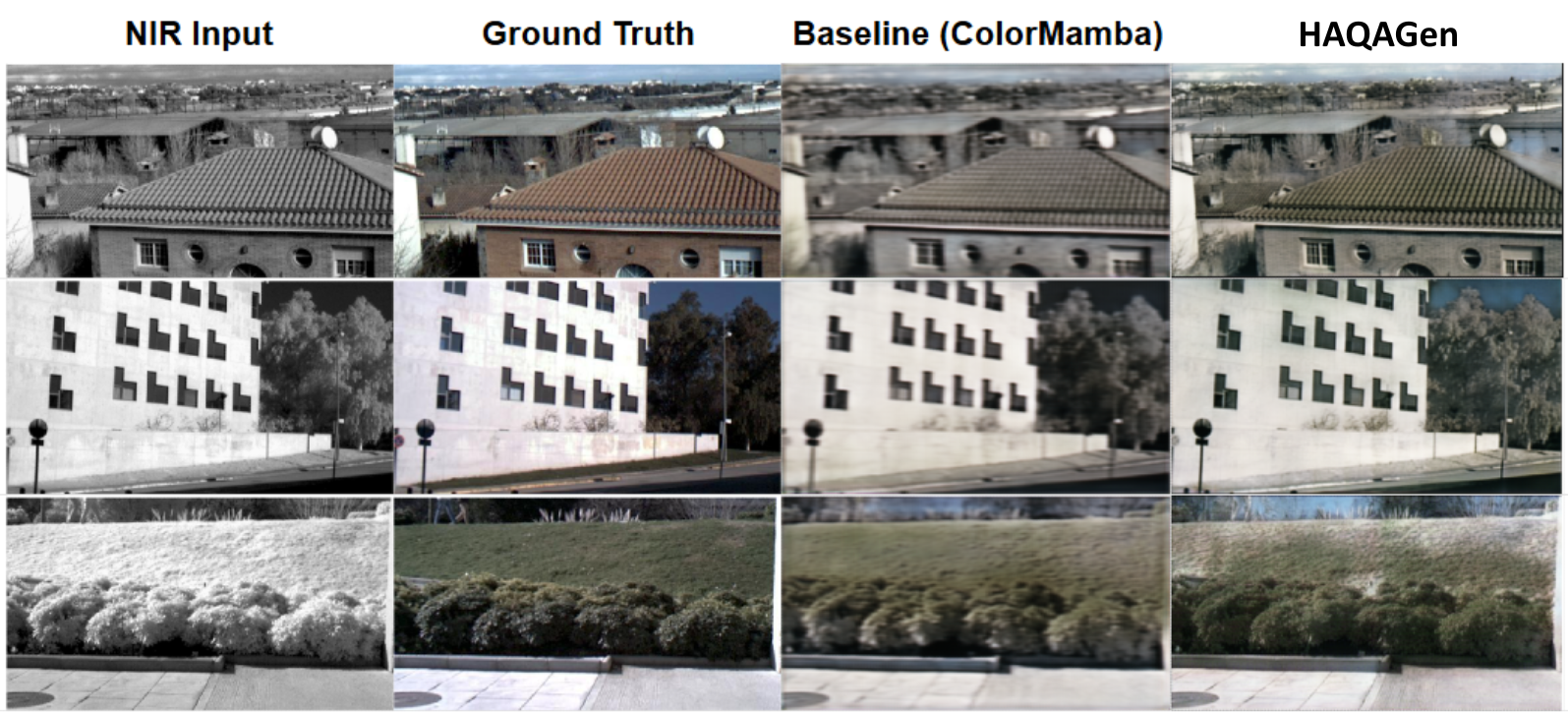}
    \caption{OMSIV~\cite{soria2017rgb-nirDataset}: Col.\,1 NIR; Col.\,2 GT; Col.\,3 ColorMamba (resized); Col.\,4 \textsc{HAQAGen} (adaptive). Sliding-window inference preserves texture and tone continuity in high-resolution settings, outperforming global resizing.}
    \label{fig:omsiv_resize}
\end{figure*}

\smallskip
\noindent \textbf{Cross-Dataset Generalization \& Adaptive Resolution }
We study generalization across FANVID, OMSIV, VCIP2020, and RGB2NIR using fixed-size vs.\ adaptive sliding-window inference. Fig.~\ref{fig:omsiv_resize} illustrates that patch-wise inference better preserves texture and tonal continuity on high-resolution imagery. Quantitatively (Table~\ref{tab:patch_vs_resize}), adaptive inference delivers consistent LPIPS and AE gains on FANVID/OMSIV/RGB2NIR. On VCIP2020, where the target resolution matches the training crop, global resizing slightly favours PSNR (consistent with reduced blending overhead), yet \textsc{HAQAGen} still achieves the best LPIPS.

\begin{table}[h]
    \centering 
    \small
    \caption{Cross-dataset comparison of \textsc{HAQAGen} vs.\ ColorMamba. Best per metric in \textbf{bold}.}
    \label{tab:dataset_comparison}
    \scalebox{0.85}{
        \begin{tabular}{l l c c c c}
            \hline
                Dataset & Method & PSNR $\uparrow$ & SSIM $\uparrow$ & AE $\downarrow$ & LPIPS $\downarrow$ \\
            \hline
                FANVID\cite{fanvid} & ColorMamba & 17.63 & 0.65 &  26.79 & 0.64\\
        \rowcolor{lightgray}  FANVID\cite{fanvid} & \textsc{HAQAGen} & \textbf{18.4} & \textbf{0.724} & \textbf{4.65} & \textbf{0.52}\\
                \hline
                OMSIV \cite{soria2017rgb-nirDataset} & ColorMamba & \textbf{17.61} & 0.58 & 25.87 &0.52\\
        \rowcolor{lightgray}  OMSIV\cite{soria2017rgb-nirDataset} & \textsc{HAQAGen} & 16.67 & \textbf{0.61} & \textbf{6.90}& \textbf{0.37} \\
                \hline
                VCIP2020 \cite{Yang2023aCoColor} & ColorMamba & 24.56 & 0.71 & \textbf{ 2.81} & 0.21 \\
        \rowcolor{lightgray}      VCIP2020\cite{Yang2023aCoColor} & \textsc{HAQAGen} & \textbf{24.96} & \textbf{0.71}& 2.96 & \textbf{0.18} \\
                \hline
                RGB2NIR \cite{BS11} & ColorMamba & \textbf{17.22} & 0.58 & 29.30 & 0.61 \\
        \rowcolor{lightgray}   RGB2NIR \cite{BS11}  & \textsc{HAQAGen} & 15.97 & \textbf{0.60} & \textbf{7.41} &\textbf{0.38}\\
        \hline
        \end{tabular}}
    \label{tab:patch_vs_resize}
\end{table}

\smallskip
\label{subsec:qualitative}
\noindent \textbf{Qualitative Analysis }
Fig.~\ref{fig:qualitative} contrasts ColorMamba with \textsc{HAQAGen} (using $\mathcal{L}_{\text{rec}}$). We consistently observe:
\begin{enumerate}
  \item \textbf{Texture Fidelity}: Fine details (foliage, contours, fabric) are better preserved, with reduced oversmoothing relative to adversarial-only baselines.
  \item \textbf{Chromatic Realism}: The CDF prior curbs tinting and enforces natural tonal distributions across materials.
  \item \textbf{Edge Consistency}: Boundaries at depth changes remain aligned after colourization, suggesting SPADE-conditioned decoding improves local hue assignment. Similar behaviour is visible at larger scales.
\end{enumerate}
\medskip
\label{sec:ablation}
\noindent \textbf{Ablation Studies}
Table~\ref{tab:loss_ablation} evaluates reconstruction-loss variants on VCIP2020. Replacing the baseline objective (MSE+Cosine+SSIM) with \emph{VGG Perceptual} alone reduces PSNR by 1.12\,dB and nearly doubles AE, indicating that perceptual loss without statistics/texture guidance is insufficient. \emph{Histogram-only} narrows AE to 3.66 but lacks sharpness (SSIM 0.68). Our composite $\mathcal{L}_{\text{rec}}$ provides the best PSNR while balancing AE and SSIM, confirming the complementarity of texture features (AE, cosine) and global statistics (CDF).
The extended ablation in the second table corroborates that removing either the CDF term or the AE-based texture supervision degrades colour or structure, respectively.

\begin{table}[h]
    \centering \small
        \begin{tabular}{lccc}
            \toprule
            \textbf{Loss Variant} & \textbf{PSNR$\uparrow$} & \textbf{SSIM$\uparrow$} & \textbf{AE$\downarrow$}\\
            \midrule
            MSE + Cosine (ColorMamba) & 24.56 & 0.71 & 2.81\\
            \quad + VGG perceptual & 23.63 & 0.70 & 4.32\\
            \quad + Histogram only & 23.81 & 0.68 & 3.66\\
            \quad + Texture ($f$) only & 24.12 & 0.69 & 3.01\\
           \rowcolor{lightgray} \textbf{Full $L_{\text{rec}}$ (ours)} & \textbf{24.96} & \textbf{0.71} & \textbf{2.96}\\
            \bottomrule
        \end{tabular}  
        \caption{Ablation on reconstruction losses (VCIP2020). Composite $L_{rec}$ balances fidelity, color, and structure.}
        \label{tab:loss_ablation}
\end{table}

\begin{table}[h]
\centering
\caption{Ablation on the HSV-SPADE branch. Removing HSV-SPADE conditioning degrades AE and SSIM, confirming that spatial hue priors improve local chromatic consistency.}
\scalebox{0.95}{\begin{tabular}{lccc}
\toprule
\textbf{Variant} & \textbf{PSNR} $\uparrow$ & \textbf{SSIM} $\uparrow$ & \textbf{AE} $\downarrow$ \\
\midrule
Without HSV-SPADE branch          & 24.21          & 0.69          & 3.52 \\
\rowcolor{lightgray} With HSV-SPADE (Ours) & \textbf{24.96} & \textbf{0.71} & \textbf{2.96} \\
\bottomrule
\end{tabular}}
\label{tab:hsv_spade_ablation}
\end{table}

\noindent \textbf{Qualitative Gallery } Resizing predictions to the original resolution introduces blur and geometric distortion, especially on FANVID and OMSIV. Adaptive patching avoids this by predicting at native scale, preserving edges and micro-texture.
\smallskip

\noindent Beyond quantitative metrics and ablation results, it is important to emphasize the broader implications of HAQAGen’s performance. The improvements in perceptual quality (LPIPS), chromatic fidelity (AE), and structural preservation (SSIM) are not merely incremental gains but represent a step toward bridging the gap between synthetic translation and real-world usability. The qualitative comparisons (Figures~\ref{fig:qualitative}--\ref{fig:omsiv_patch_vs_resize}) consistently show that HAQAGen avoids the common pitfalls of over-smoothing and spectral ambiguity that plague prior methods. Instead, it generates outputs that maintain edge sharpness and textural richness, qualities that directly impact downstream tasks such as detection and recognition. The robustness across datasets of varying resolution and content diversity further highlights the scalability of the framework. Collectively, these findings validate HAQAGen not only as a strong benchmark model for NIR-to-RGB translation but also as a practical tool for real-world deployment where both perceptual realism and structural fidelity are paramount.

\section{Conclusion}
\label{sec:conclusion}

In this paper, we introduced \textbf{HAQAGen}, a unified histogram-assisted framework that advances the frontier of NIR-to-RGB spectral translation. By jointly leveraging global colour statistics, HSV-based chromatic priors, and texture-aware feature supervision within a Mamba backbone, HAQAGen resolves the trade-off between \emph{chromatic realism} and \emph{textural fidelity}. Extensive experiments across four diverse benchmarks demonstrated noticeable improvement: quantitatively, HAQAGen achieves gains of up to \textbf{1.63\,dB} in PSNR and \textbf{15.09\%} improvement in LPIPS over state-of-the-art methods; qualitatively, it produces outputs with vivid colours, sharp structural boundaries, and reliable preservation of scene detail across varying scales and environments. Moreover, the adaptive-resolution inference engine ensures scalability to high-resolution imagery, enabling real-time deployment on commodity hardware without sacrificing quality.
\smallskip

Beyond numerical performance, our analyses highlight HAQAGen’s practical impact. Its robustness across disparate datasets and strong compatibility with downstream tasks (e.g., object detection) indicate that NIR-to-RGB translation can evolve from a purely generative challenge to a foundation for actionable perception in adverse visual conditions. 
\smallskip

Looking forward, several directions hold promise: (i) exploring \emph{self-supervised colour priors} to reduce reliance on paired RGB supervision, (ii) distilling the dual-branch architecture into ultra-lightweight variants tailored for edge devices, and (iii) joint optimisation with higher-level tasks such as segmentation, tracking, and low-light enhancement to enable end-to-end NIR-aware vision systems. 
\smallskip

We believe HAQAGen establishes a strong step toward practical and scalable NIR-to-RGB colourisation, laying the groundwork for next-generation perception in autonomous systems, security, remote sensing, and other human-centric applications where visibility is mission-critical.
\clearpage
{
    \small
    \bibliographystyle{ieeenat_fullname}
    \bibliography{our}
}

\end{document}